\documentclass{article}
\usepackage{spconf,amsmath,epsfig}
\usepackage{color}
\usepackage{booktabs}
\usepackage{paralist}

\usepackage{subfig}
\usepackage{graphicx}
\usepackage{url}
\usepackage[font=normal,skip=7pt]{caption}%reduce spacing between figure and figure captions slightly from default of 10pt

\graphicspath{{figures/}}

\title{CUMULATIVE ASSESSMENT FOR URBAN 3D MODELING}
\name{Shea Hagstrom$^1$, Hee Won Pak$^1$, Stephanie Ku$^1$, Sean Wang$^1$, Gregory Hager$^2$, and Myron Brown$^1$}
\address{$^1$JHU Applied Physics Laboratory \quad $^2$The Johns Hopkins University}

\begin{document}
\maketitle

% Abstract
\begin{abstract}
Urban 3D modeling from satellite images requires accurate semantic segmentation to delineate urban features, multiple view stereo for 3D reconstruction of surface heights, and 3D model fitting to produce compact models with accurate surface slopes. In this work, we present a cumulative assessment metric that succinctly captures error contributions from each of these components. We demonstrate our approach by providing challenging public datasets and extending two open source projects to provide an end-to-end 3D modeling baseline solution to stimulate further research and evaluation with a public leaderboard.
\end{abstract}

% Index terms

% Introduction
\section{Introduction}
\label{sec:intro}

Urban 3D models play an important role in a variety of applications such as city planning, tourism, navigation, and emergency management. Automated methods for producing these models from multiple satellite images are especially important for use cases involving coverage of large geographic areas or remote geographic regions inaccessible by airborne cameras or lidar.

The component challenges for urban 3D geometric modeling include semantic segmentation to delineate urban features, multiple view stereo (MVS) for 3D reconstruction of surface heights, and 3D model fitting to produce compact mesh models with accurate roof slopes. Comprehensive evaluation must account for error contributions from each to enforce model quality. For example, Rottensteiner et al. \cite{ROTTENSTEINER2014256} assessed semantic, geometric, and topological error for 3D building reconstruction from airborne images. Similarly, Bosch et al. \cite{Bosch2017METRICEP} assessed semantic, volumetric, and curvature error for building modeling from satellite images. Both derived reference values for evaluation from airborne lidar and assessed each metric separately. To the best of our knowledge, there has been no published work to define a holistic metric for end-to-end evaluation accounting for all error contributions with a single value.

Li et al. \cite{8899241} and Leotta et al. \cite{Leotta_2019_CVPR_Workshops} have recently proposed end-to-end pipelines for 3D building reconstruction from satellite images. The Danesfield pipeline \cite{Leotta_2019_CVPR_Workshops} is open source except for a commercial MVS algorithm, while Li et al.'s pipeline \cite{8899241} employs open source S2P \cite{s2p8014932} for MVS but is otherwise not public. We are not aware of any end-to-end pipeline that is entirely open source. 

The ISPRS benchmark \cite{ROTTENSTEINER2014256} provides public data for assessing urban 3D building reconstruction from airborne images. While public data is widely available individually for semantic segmentation \cite{Etten2018SpaceNetAR,demir2018deepglobe}, MVS \cite{bosch2016mvs3dm,Sonali2019}, and semantic stereo \cite{bosch2019semantic,j9229514}, we are not aware of any public datasets for evaluation of end-to-end building modeling pipelines with satellite images. Notably, reference data used for evaluations reported in \cite{8899241} and \cite{Leotta_2019_CVPR_Workshops} were not previously publicly available.

\begin{figure}
  \captionsetup[subfigure]{labelformat=empty}
  \hfill
  \subfloat[Semantics]{\includegraphics[width=0.32\columnwidth]{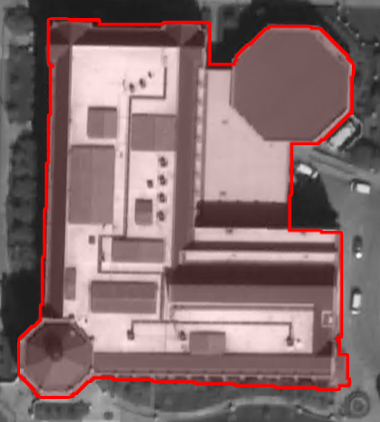}}
  \hfill
  \subfloat[Elevation]{\includegraphics[width=0.32\columnwidth]{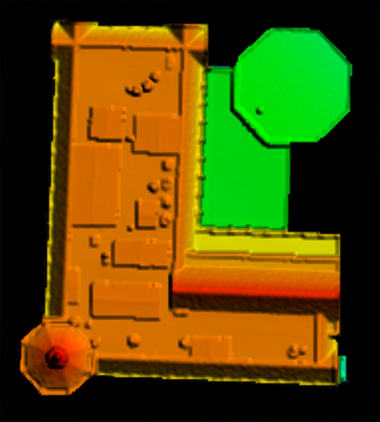}}
  \hfill
  \subfloat[Slope angles]{\includegraphics[width=0.32\columnwidth]{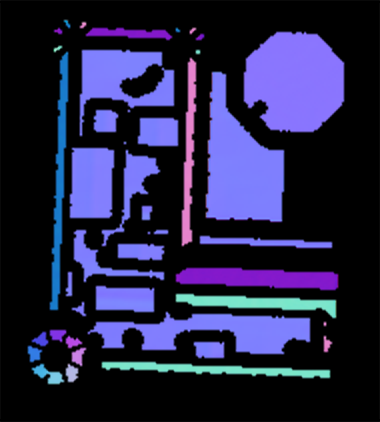}}
  \hfill\null
  \caption{Our metrics for cumulative assessment capture errors in semantic segmentation, multi-view stereo, and 3D model fitting. Image is courtesy DigitalGlobe.}
  \label{fig:teaser}
\end{figure}

We begin to address these observed limits of prior work with the following contributions:
\begin{compactitem}
\item We propose a cumulative assessment metric for urban 3D modeling from overhead images that explicitly captures error contributions from semantic segmentation, multiple view stereo, and 3D model fitting based on elevation and surface orientations (Fig. \ref{fig:teaser}).
\item To provide an end-to-end open source baseline solution for public evaluation, we extend the Danesfield pipeline \cite{Leotta_2019_CVPR_Workshops} with open source VisSat MVS \cite{VisSat19} to replace the commercial MVS component which is not publicly available.
\item We provide challenging test sets and evaluate both Danesfield \cite{Leotta_2019_CVPR_Workshops} and our extended baseline solution. To stimulate further research, we make our code and test sets publicly available \cite{pubgeo_core3d_open} with an online leaderboard for evaluation.
\end{compactitem}

% Methods
\section{Methods}
\label{sec:methods}

\subsection{Cumulative Assessment}
\label{ssec:cumulative-assessment}
Standard 2D Intersection Over Union (IOU) evaluation using semantic labels is our starting point for making a single assessment metric incorporating both labeling and modeling. 3D mesh models are converted to rasterized Digital Surface Models (DSMs) for the purpose of evaluating geometry. Rasterization limits our evaluation to 2.5D simplifications, though in practice we prefer this for the simplicity in large-scale evaluation. Prior to evaluation, the test model is registered to our ground truth to remove any translation offset.
IOU for 2D semantic labels is
\begin{align*}
\mathit{TP}_c=\sum\nolimits_{i}^{N}\mathit{TP}_c^i \ \ \mathrm{where} \ \mathit{TP}_c^i=1 \ \mathit{iff} \ \hat{C}_i=1 \ \mathrm{and} \ C_i=1\\
\mathit{FP}_c=\sum\nolimits_{i}^{N}\mathit{FP}_c^i \ \ \mathrm{where} \ \mathit{FP}_c^i=1 \ \mathit{iff} \ \hat{C}_i=1 \ \mathrm{and} \ C_i=0\\
\mathit{FN}_c=\sum\nolimits_{i}^{N}\mathit{FN}_c^i \ \ \mathrm{where} \ \mathit{FN}_c^i=1 \ \mathit{iff} \ \hat{C}_i=0 \ \mathrm{and} \ C_i=1
\end{align*}
\begin{equation}
\mathit{IOU}_c=\frac{\mathit{TP}_c}{\mathit{TP}_c + \mathit{FN}_c + \mathit{FP}_c}
\end{equation}
where each pixel has a known binary label for category $C$ in addition to predicted category $\hat{C}$, and we define true positive ($\mathit{TP}$), false positive ($\mathit{FP}$), and false negative ($\mathit{FN}$) binary counts from these.
An additional binary label can be derived for error in elevation Z by accepting only values where the Z error is less than a defined threshold, as in Equation \eqref{eqn:ZThreshold}.
The Z constraint helps ensure building model height and volume are correct, but it does not account for the correctness of localized slopes. For this purpose we introduce the slope error, defined as the angle between the reference and test model surface normals as in Equation \eqref{eqn:AngleThreshold}. Local normals using Z values within a few meters of each pixel are computed from both for comparison using eigendecomposition. While this is well-defined for planar regions, at sharp edges or complex geometry the reference normals become more ambiguous. We use the ratios of the three eigenvalues to exclude these unstable reference normals and use only values passing the constraints
\begin{align*}
\lambda_3/\left(\lambda_1+\lambda_2+\lambda_3\right)&<0.005\\
\left(\lambda_2-\lambda_3\right)/\lambda_1&>0.2
\end{align*}
In our tests we found reference pointing normal errors to be under 3 degrees in 90\% of samples, and Fig. \ref{fig:roof-slopes} shows that errors are often even lower. The slope plot in Fig. \ref{fig:teaser} shows the regions with sufficient planarity to use in evaluation and excludes edges and small features.

The $TP$ binary semantic values are pixel-wise multiplied by the corresponding binarized Z or angle errors, each successively reducing the $TP$ counts. The full equation becomes
\begin{equation}\label{eqn:ZThreshold}
\mathit{TP_z^i}=1 \ \mathit{iff} \ |\hat{Z_i}-Z_i|<\mathrm{1.0 \ meter}
\end{equation}
\begin{equation}\label{eqn:AngleThreshold}
%\mathit{TP_\theta^i}=1 \ \mathit{iff} \ |\hat{\theta_i}-\theta_i|<\mathrm{5.0 \ degrees}\\
\mathit{TP_\theta^i}=1 \ \mathit{iff} \ \mathrm{cos}^{-1}(\hat{\vec{u_{i}}}\cdot\vec{u_{i}}) <\mathrm{5.0 \ degrees}\\
\end{equation}
\begin{equation}
\mathit{TP_m}=\sum\nolimits_i^N \mathit{TP_c^i}\mathit{TP_z^i}\mathit{TP_\theta^i}
\end{equation}
\begin{equation}
\mathit{IOU_m}=\frac{\mathit{TP_m}}{\mathit{TP_c} + \mathit{FN_c} + \mathit{FP_c}}
\end{equation}
We also define an additional $\mathit{IOU_z}$ as an intermediate product not incorporating slope information. This gives a cumulative progression from $\mathit{IOU_c}$ to $\mathit{IOU_z}$ to $\mathit{IOU_m}$ where each incorporates an additional constraint on the model accuracy.

In practice, both Z error and orientation error are not evaluable at every sample due to lack of valid reference values. In these cases we have chosen to assume that the respective measurements are correct in the IOU totals. Another approach would be to only evaluate models where all reference values are valid, however this would reduce the total area evaluated, in some cases significantly.

\subsection{Public Baseline}
\label{ssec:public-baseline}

To provide an open source baseline solution, we extended the Danesfield pipeline \cite{Leotta_2019_CVPR_Workshops}. Danesfield performs semantic segmentation to delineate buildings, with inputs from multi-spectral image channels, normalized difference vegetation index (NDVI) values computed from those channels, and a normalized digital surface model (nDSM). The nDSM is derived by first segmenting ground and producing a digital terrain model (DTM) and then subtracting the DTM from the MVS digital surface model (DSM) to get height above ground. The MVS solution in Danesfield is commercial and not available publicly. MVS point clouds for segmented roofs are further segmented to produce roof shapes that are merged to produce building models.

We replaced the commercial MVS solution with open source VisSat \cite{VisSat19}, an adaptation of the state-of-the-art COLMAP MVS pipeline \cite{schoenberger2016sfm} for use with satellite images. Point clouds were initially noisier than those produced by the commercial MVS solution, so we added a point cloud bilateral filter \cite{ipol.2017.179} in post-processing. Details for reproducing our results are available in \cite{pubgeo_core3d_open}.

% Experiments
\section{Experiments}
\label{sec:experiments}

%% Dataset
\subsection{Dataset}

\begin{figure}[t]
	\includegraphics[width=\columnwidth]{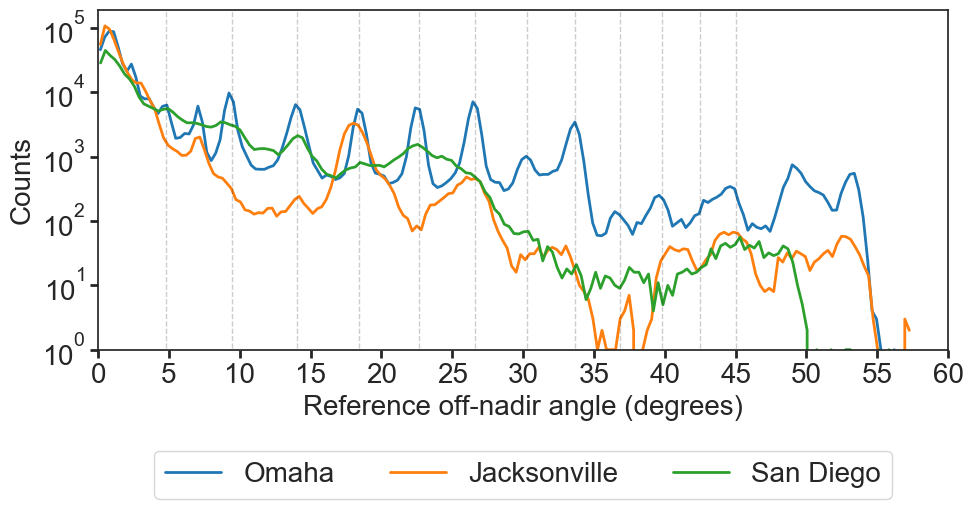}
	\caption{Distribution histograms of reference roof angles from flat (0 degrees) are shown for our three test sites. Vertical lines represent roof pitches from 1/12 to 12/12 in 1/12 increments. Counts are plotted with a log scale.}
	\label{fig:roof-slopes}
\end{figure}

Our test set consists of multiple DigitalGlobe WorldView-3 satellite images over areas of interest (AOIs) from Jacksonville (2 sq. km), San Diego (1 sq. km), and Omaha (1 sq. km) \cite{pubgeo_core3d_open} (Fig. \ref{fig:textured-models}). Jacksonville and San Diego AOIs were also evaluated in \cite{8899241,Leotta_2019_CVPR_Workshops}. These AOIs are well-suited for semantic segmentation using available public training datasets, and each has dozens of images suitable for accurate MVS. AOIs also vary in modeling difficulty, with Jacksonville roofs being mostly large and flat and Omaha and San Diego having much more variety in roof complexity and pitch (Fig. \ref{fig:roof-slopes}). Omaha in particular has a variety of standard roof pitches in this region, made apparent by the many histogram peaks. Evaluation is conducted with manually edited building semantic labels with height and slope derived from airborne lidar.

%% Results
\subsection{Results}
In Tab. \ref{tab:results}, we compare metrics for Danesfield 3D reconstruction accuracy using open source VisSat MVS \cite{VisSat19} and using commercial MVS from \cite{Leotta_2019_CVPR_Workshops}. Cumulative metrics from section \ref{ssec:cumulative-assessment} are reported along with root mean square (RMS) error statistics for roof height and slope. For both pipelines, we assessed building models produced directly from MVS 3D points and also models produced after fitting those points to roof primitives. Improvement in $\mathit{IOU_m}$ and $\mathit{RMS_\theta}$ values clearly indicate the efficacy of model fitting for achieving more accurate roof slopes. Upon inspection, we see that flat roofs are modeled very well and sloped roofs much less so, accounting for increased $\mathit{RMS_z}$ values. To achieve a high $\mathit{IOU_m}$ score, semantic segmentation, MVS, and 3D model fitting algorithms must all perform well. We hope that our work will encourage more attention to each, but particularly model fitting which has received less attention.

Our filtered VisSat MVS point clouds achieve consistently lower $\mathit{RMS_z}$ and $\mathit{RMS_\theta}$ values than the commercial MVS solution from \cite{Leotta_2019_CVPR_Workshops}; however, relative horizontal accuracy is inferior for the San Diego test site, leading to less accurate ground segmentation which contributes to error in building segmentation and thus lower $\mathit{IOU_c}$ values. Building segmentation methods relying only on RGB images and trained using public data have been shown to score higher for IOU-2D on our test sites. For example, \cite{8899241} reports 0.86 for Jacksonville and 0.82 for San Diego. We expect our baseline can be improved by replacing the semantic segmentation module with a high-performing solution from the SpaceNet competition \cite{Etten2018SpaceNetAR}.

Textured building models for the three AOIs are shown in Fig. \ref{fig:textured-models} along with photos of the full modeled areas. An example of cumulative error statistics is shown in Fig. \ref{fig:jax-metrics-example} for a building in Jacksonville. Semantic segmentation errors are apparent, such as trees mislabeled as building shown in red. The MVS DSM has accurate height but incorrect slopes on the flat roof. This is much improved after successful model fitting; however, modeling with the baseline is less successful for complex and pitched roofs, as shown in Tab. \ref{tab:results}. We hope that the cumulative metrics we have presented will inspire research to more robustly address these challenges.

\begin{table}[h]
	\setlength{\tabcolsep}{5pt}
	\resizebox{\columnwidth}{!}{
		\begin{tabular}{l|ccc|cc}  
			\toprule  
			& $\mathit{IOU_c}$ & $\mathit{IOU_z}$ & $\mathit{IOU_m}$ & $\mathit{RMS_z}$ & $\mathit{RMS_\theta}$ \\ 
			\midrule
            $\textbf{JACKSONVILLE}$ & & & & & \\
			Commercial MVS & 0.63 & 0.52 & 0.33 & 0.72 & 13.39 \\
			Commercial MVS + Modeling & 0.63 & 0.51 & 0.44 & 1.39 & 6.99 \\
			VisSat MVS & 0.67 & $\textbf{0.56}$ & 0.41 & $\textbf{0.50}$ & 8.80 \\ 
			VisSat MVS + Modeling & $\textbf{0.68}$ & $\textbf{0.56}$ & $\textbf{0.50}$ & 0.71 & $\textbf{3.53}$ \\ 
			\midrule
            $\textbf{SAN DIEGO}$ & & & & & \\
			Commercial MVS & $\textbf{0.67}$ & $\textbf{0.53}$ & 0.32 & 0.86 & 17.77 \\
			Commercial MVS + Modeling & $\textbf{0.67}$ & 0.52 & $\textbf{0.40}$ & 1.88 & 15.14 \\
			VisSat MVS & 0.58 & 0.47 & 0.32 & $\textbf{0.70}$ & 15.89 \\ 
			VisSat MVS + Modeling & 0.59 & 0.47 & 0.38 & 0.99 & $\textbf{10.32}$ \\ 			
			\midrule
            $\textbf{OMAHA}$ & & & & & \\
			VisSat MVS & $\textbf{0.58}$ & $\textbf{0.52}$ & 0.34 & $\textbf{0.37}$ & 9.32 \\ 
			VisSat MVS + Modeling & $\textbf{0.58}$ & 0.50 & $\textbf{0.40}$ & 0.57 & $\textbf{8.35}$ \\ 
			\bottomrule
		\end{tabular}
	}
	\caption{$\mathit{IOU_m}$ and $\mathit{RMS_\theta}$ metrics for Danesfield 3D reconstruction accuracy using open source VisSat MVS \cite{VisSat19} and using commercial MVS from \cite{Leotta_2019_CVPR_Workshops} show the value of roof modeling for improving 3D reconstruction.}
	\label{tab:results}
\end{table}

\newlength{\imht}
\begin{figure}[]
\setlength{\imht}{0.21\columnwidth}%0.23 is widest possible
\footnotesize
\centering
\begin{tabular}{@{}c@{\hspace{0.1in}}@{}c@{}}
\includegraphics[height=\imht]{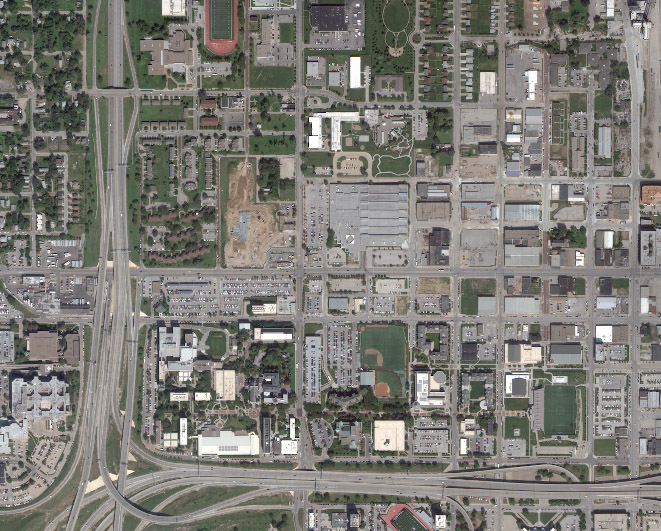} & \includegraphics[height=\imht]{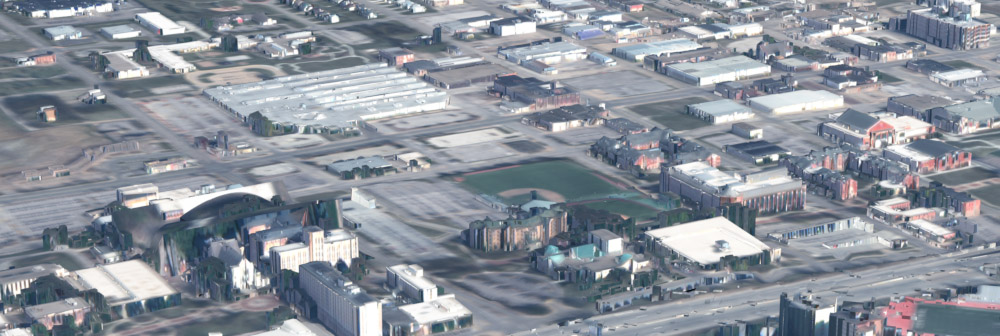}\\
Omaha AOI & Omaha mesh model\\[0.06in]
\includegraphics[height=\imht]{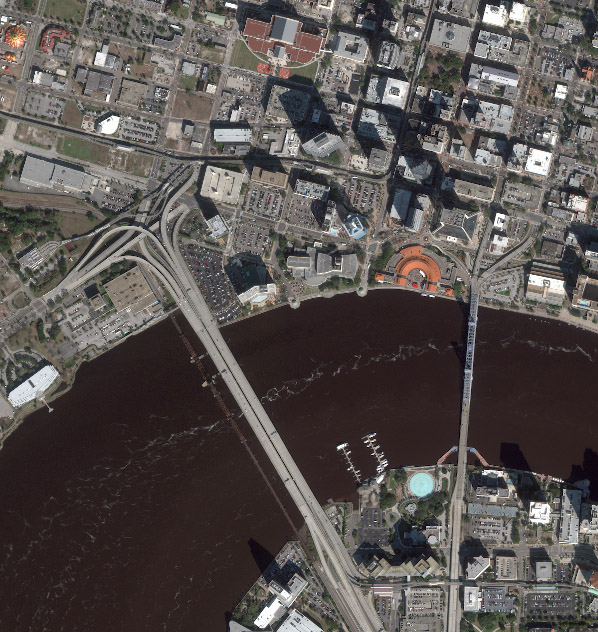} & \includegraphics[height=\imht]{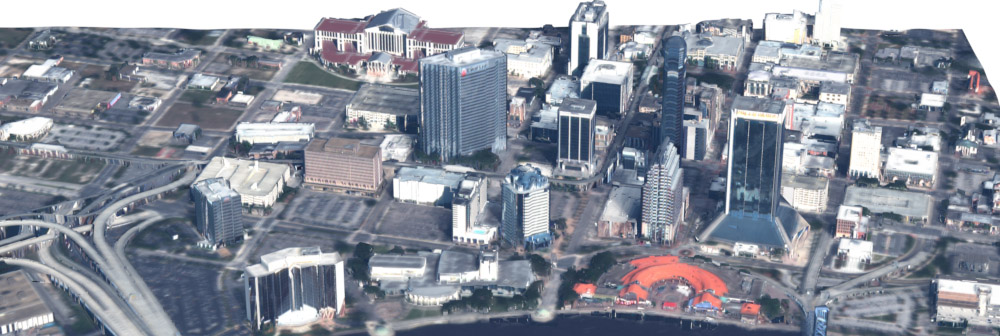}\\
Jacksonville AOI & Jacksonville mesh model\\[0.06in]
\includegraphics[height=\imht]{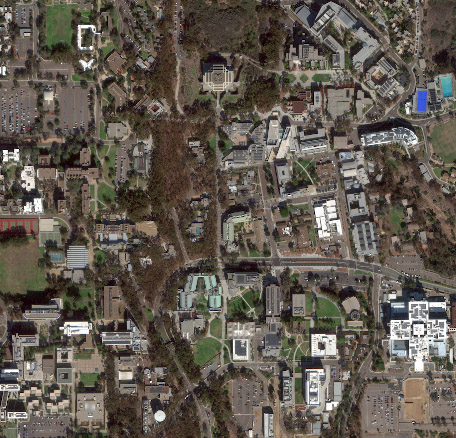} & \includegraphics[height=\imht]{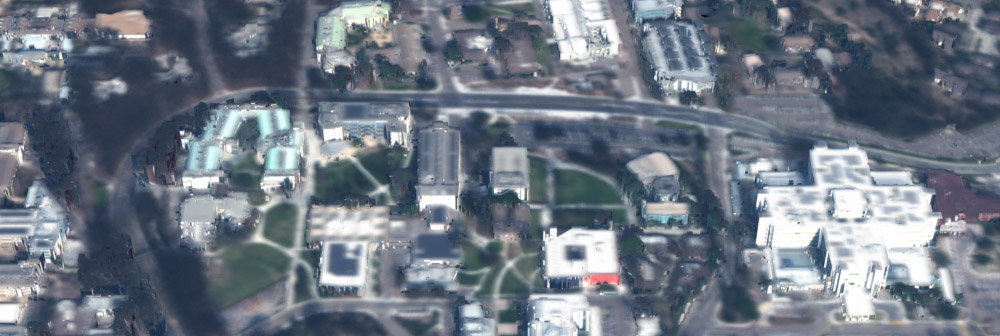}\\
San Diego AOI & San Diego mesh model
\end{tabular}
\caption{AOI images and renders of textured building models from Danesfield / VisSat. Images are courtesy DigitalGlobe.}
\label{fig:textured-models}
\end{figure}

\setlength{\imht}{0.7in}
\begin{figure}[h]
    \captionsetup[subfigure]{labelformat=empty}
    \captionsetup[subfloat]{farskip=4pt,captionskip=2pt}%farskip=distance between subs, captionskip=distance between subfig and subfig caption
    \hfill
    \subfloat[Google Earth RGB]{\makebox[1.5\width]{\includegraphics[height=\imht]{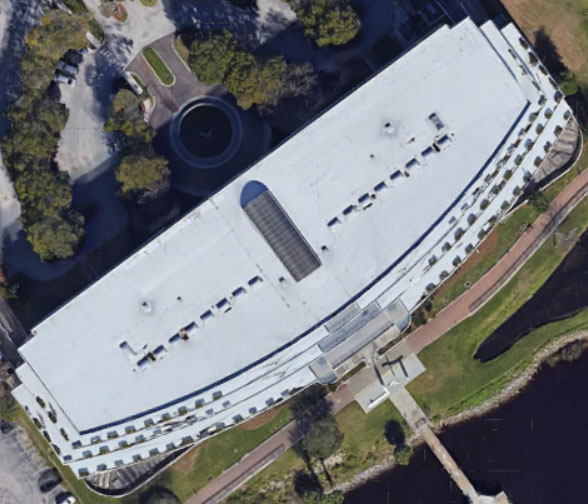}}}
    \hfill
    \subfloat[Reference lidar]{\makebox[1.5\width]{\includegraphics[height=\imht]{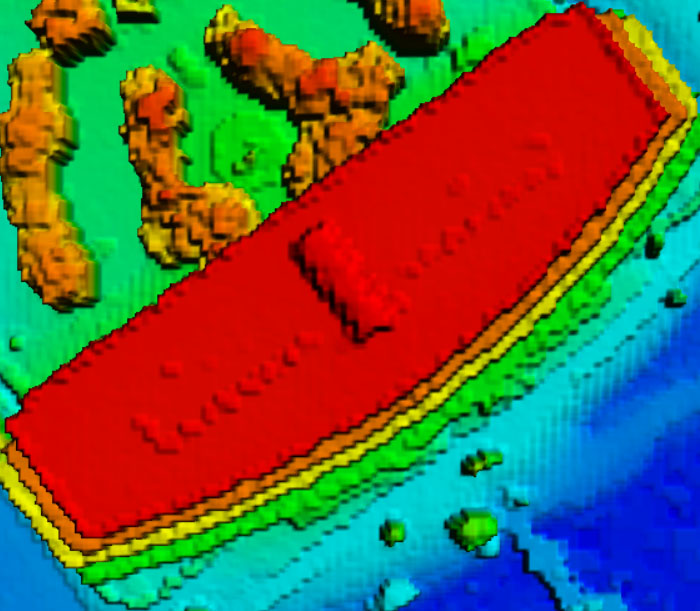}}}
    \hfill\null\\
    \subfloat[MVS DSM]{\includegraphics[height=\imht]{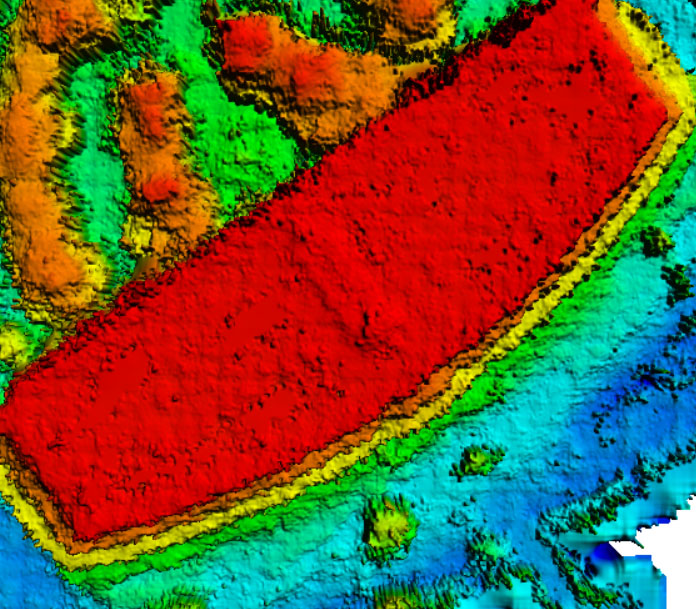}}
    \hfill
    \subfloat[Semantic]{\includegraphics[height=\imht]{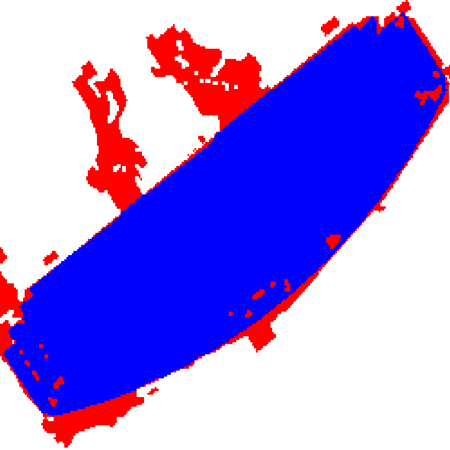}}
    \hfill
    \subfloat[+ height]{\includegraphics[height=\imht]{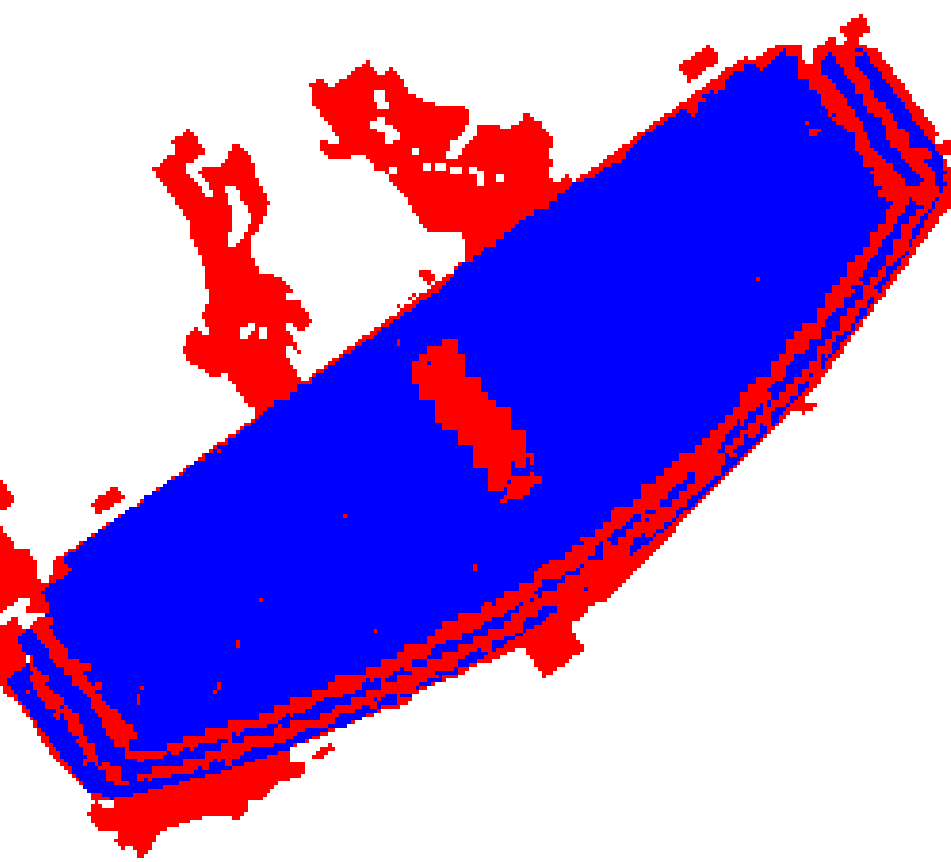}}
    \hfill
    \subfloat[+ slope]{\includegraphics[height=\imht]{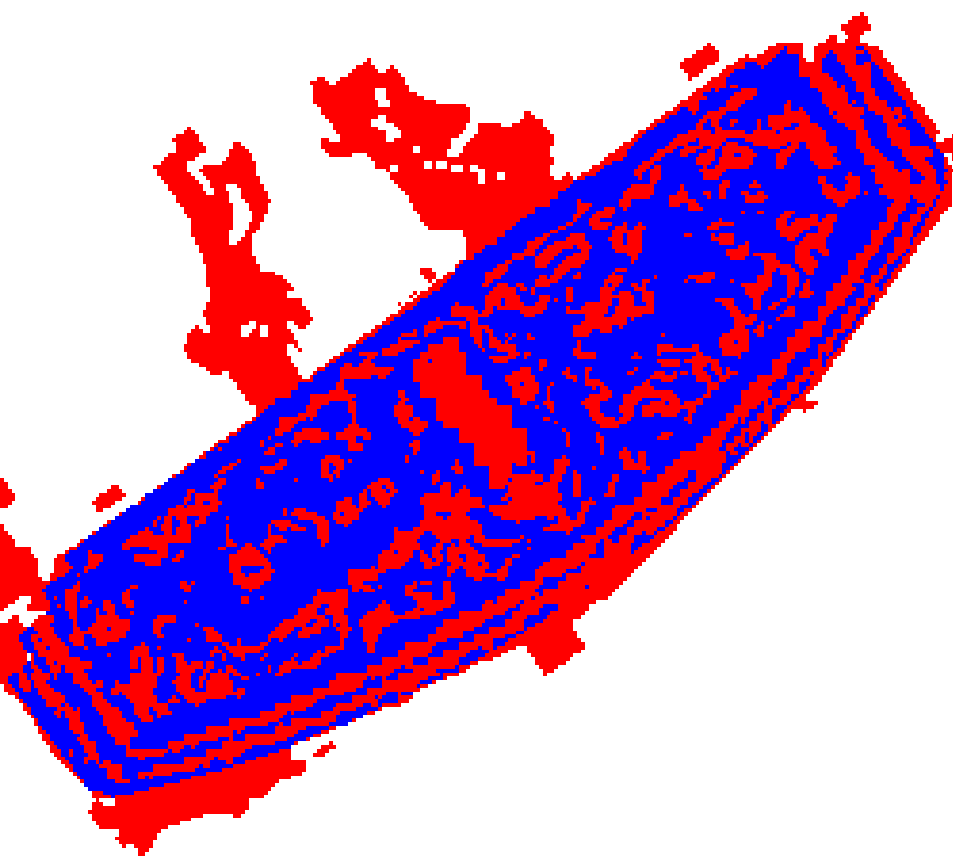}}
    \hfill\null\\
    \subfloat[3D model]{\includegraphics[height=\imht]{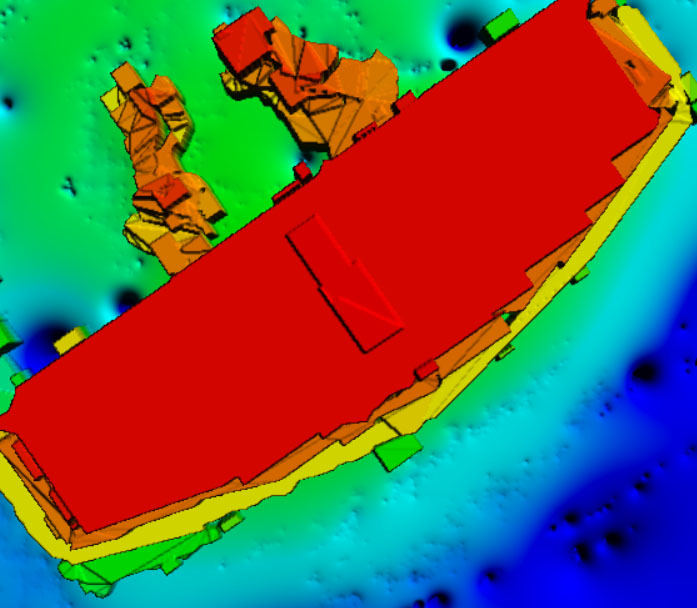}}
    \hfill
    \subfloat[Semantic]{\includegraphics[height=\imht]{pgf-sem.png}}
    \hfill
    \subfloat[+ height]{\includegraphics[height=\imht]{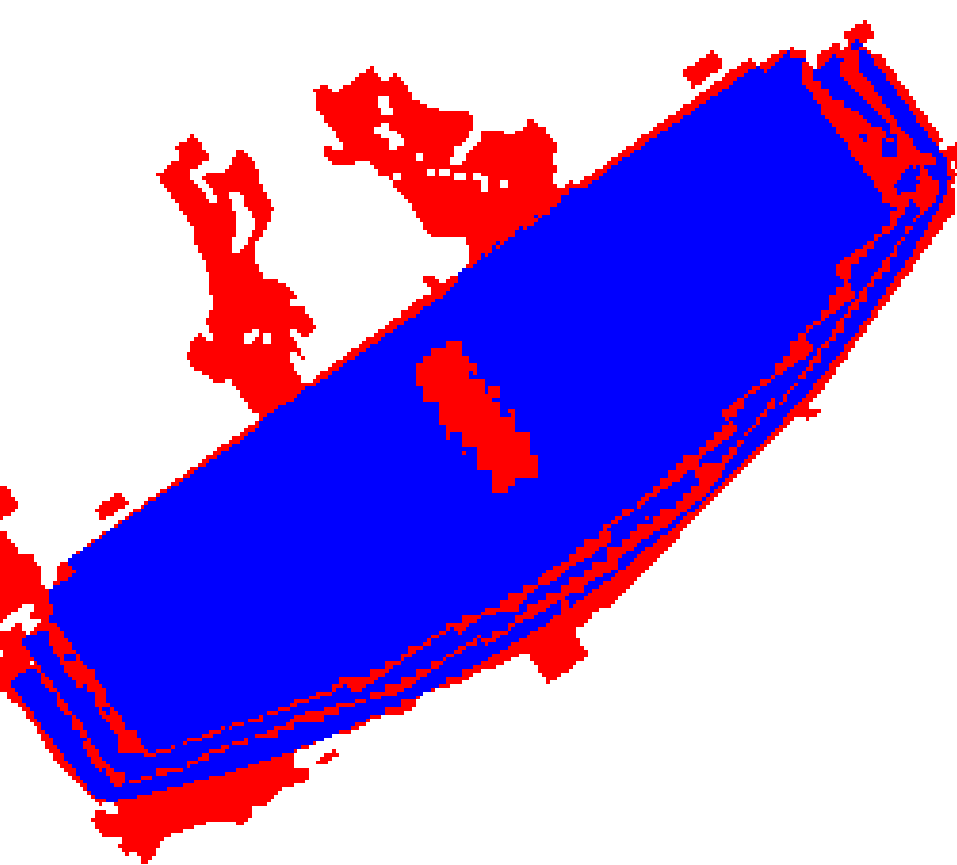}}
    \hfill
    \subfloat[+ slope]{\includegraphics[height=\imht]{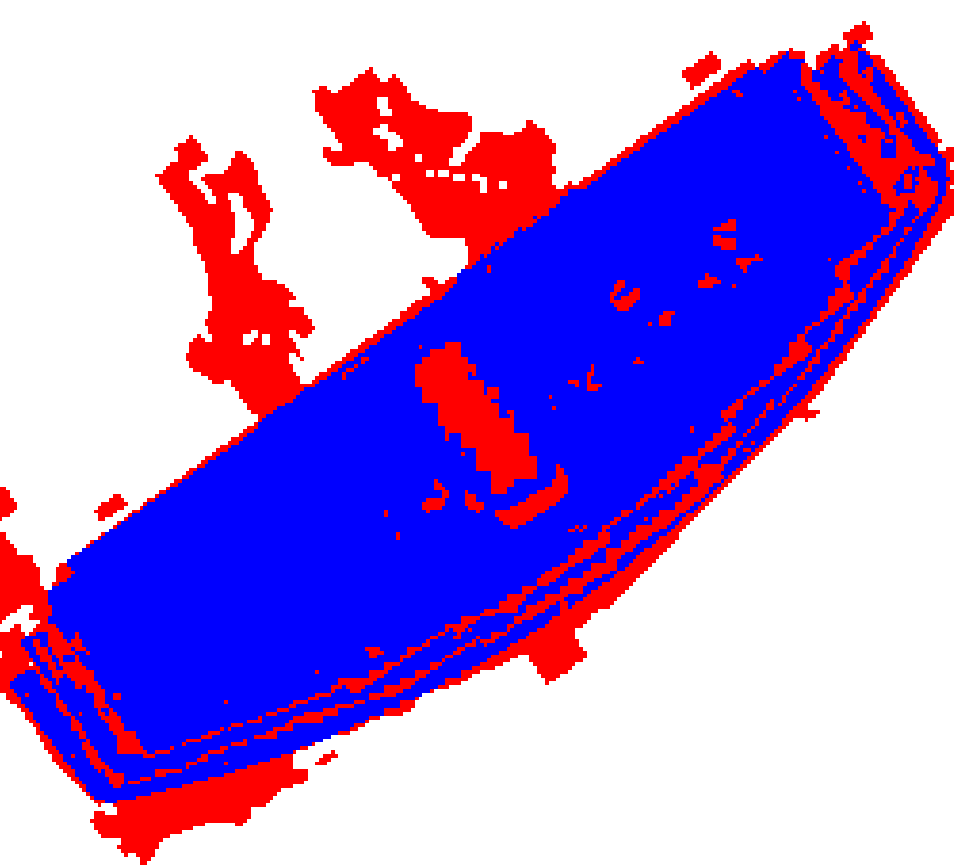}}
    \hfill\null
	\caption{Model heights from MVS and after 3D model fitting are compared. Cumulative evaluation constraints show the increasing failed areas (red) and passed areas (blue). Lidar and Google Earth RGB images are shown for reference.}
	\label{fig:jax-metrics-example}
\end{figure}

%old version
%\begin{figure}[h]
%	\includegraphics[width=\columnwidth]{jax-metrics-example.png}
%	\caption{Model heights from MVS and after 3D model fitting are compared. Areas passing all listed evaluation constraints are shown blue and those failing in red. Lidar and Google Earth RGB images are shown for reference.}
%	\label{fig:jax-metrics-example}
%\end{figure}

% Conclusion
\section{Conclusion}
\label{sec:conclusion}
We have presented cumulative assessment metrics for urban 3D modeling leading to a single score capturing error contributions from semantic segmentation, multiple view stereo, and 3D model fitting. To encourage 3D model fitting research, we produced challenging test data and combined two open source projects to provide an end-to-end 3D modeling baseline solution. We will make our code and data publicly available with an online leaderboard for evaluation.

%spacing test
%\fbox{
%\setlength\fboxsep{0pt}
%\setlength\fboxrule{0.5pt}
%\vbox to 0.7in {\vfil
%\hbox to 3cm{Some info}%
%\vfil
%}}

% Acknowledgements
\section{Acknowledgments}
\label{sec:acknowledgments}
We thank the authors of \cite{Leotta_2019_CVPR_Workshops} and \cite{VisSat19} for their assistance and advice in facilitating our extension of their open source works. This work was supported by the Intelligence Advanced Research Projects Activity (IARPA) contract no. 2017-17032700004. The U.S. Government is authorized to reproduce and distribute reprints for Governmental purposes notwithstanding any copyright annotation thereon. Disclaimer: The views and conclusions contained herein are those of the authors and should not be interpreted as necessarily representing the official policies or endorsements, either expressed or implied, of IARPA or the U.S. Government.

% References
\bibliographystyle{IEEEbib}
\bibliography{references}

\end{document}